\pdfoutput=1

\documentclass[11pt]{article}

\usepackage{acl}

\usepackage{times}
\usepackage{latexsym}
\usepackage{algorithm}
\usepackage{algorithmic}
\usepackage{graphicx}
\usepackage{multirow}
\usepackage{multicol}
\usepackage{tabularx}
\usepackage{amsmath}
\usepackage{amssymb}
\usepackage{url}
\usepackage{footnote}
\makesavenoteenv{table*}
\usepackage[T1]{fontenc}

\usepackage[utf8]{inputenc}

\usepackage{microtype}

\title{Lite Unified Modeling for Discriminative Reading Comprehension}

\author{Yilin Zhao\textsuperscript{1,2}, Hai Zhao\textsuperscript{1,2,\thanks{\ \ Corresponding author. This paper was partially supported by Key Projects of National Natural Science Foundation of China under Grants U1836222 and 61733011.}}, Libin Shen\textsuperscript{3}, Yinggong Zhao\textsuperscript{3}\\
\textsuperscript{1} Department of Computer Science and Engineering, Shanghai Jiao Tong University\\
\textsuperscript{2} Key Laboratory of Shanghai Education Commission for Intelligent Interaction\\
and Cognitive Engineering, Shanghai Jiao Tong University, Shanghai, China\\
\textsuperscript{3}Leyan Tech, Shanghai, China\\
\texttt{zhaoyilin@sjtu.edu.cn, zhaohai@cs.sjtu.edu.cn}\\
\texttt{libin@leyantech.com, ygzhao@leyantech.com}\\
}

\begin{document}
\maketitle
\begin{abstract}
As a broad and major category in machine reading comprehension (MRC), the generalized goal of discriminative MRC is answer prediction from the given materials. 
However, the focuses of various discriminative MRC tasks may be diverse enough: multi-choice MRC requires model to highlight and integrate all potential critical evidence globally; while extractive MRC focuses on higher local boundary preciseness for answer extraction.
Among previous works, there lacks a unified design with pertinence for the overall discriminative MRC tasks.
To fill in above gap, we propose a lightweight \textbf{PO}S-Enhanced \textbf{I}terative Co-Attention \textbf{Net}work (\textit{POI-Net}) as the first attempt of unified modeling with pertinence, to handle diverse discriminative MRC tasks synchronously.
Nearly without introducing more parameters, our lite unified design brings model significant improvement with both encoder and decoder components.
The evaluation results on four discriminative MRC benchmarks consistently indicate the general effectiveness and applicability of our model, and the code is available at \url{https://github.com/Yilin1111/poi-net}.
\end{abstract}

\section{Introduction}
Machine reading comprehension (MRC) as a challenging branch in NLU, has two major categories: generative MRC which emphasizes on answer generation \cite{Schwarz2018narrativeqa}, and discriminative MRC which focuses on answer prediction from given contexts \cite{baradaran2020survey}.
Among them, discriminative MRC is in great attention of researchers due to its plentiful application scenarios, such as extractive and multi-choice MRC two major subcategories.
Given a question with corresponding passage, extractive MRC asks for precise answer span extraction in passage \cite{joshi2017triviaqa, trischler2017newsqa, yang2018hotpotqa}, while multi-choice MRC requires suitable answer selection among given candidates \cite{huang2019CosmosQA, Khashabi2018MultiRC}.
Except for the only common goal shared by different discriminative MRCs, the focuses of extractive and multi-choice MRC are different to a large extent due to the diversity in the styles of predicted answers:
multi-choice MRC usually requires to highlight and integrate all potential critical information among the whole passage; while extractive MRC pays more attention to precise span boundary extraction at local level, since the rough scope of answer span can be located relatively easily, shown in Table \ref{examples}.

\begin{table}[t]
\small
\fontsize{9pt}{\baselineskip}\selectfont
\centering
\begin{tabular}{|p{7.3cm}|}
\hline \bf Multi-choice MRC Example \\\hline
\emph{... In addition, Lynn's pioneering efforts also provide public educational \textbf{forums} via \textbf{Green Scenes} -- \textbf{a series of three hour events}, each focusing on specific topics teaching Hoosiers how to lead \textbf{greener lifestyles}. ...}\\\hline
Q: \emph{What can we learn about \textbf{Green Scenes}?}\\
A. \emph{It is a scene set in a \textbf{three-hour film}.}\\
B. \emph{It is \textbf{a series of events} focusing on \textbf{green life}.} (\textbf{Golden})\\
C. \emph{It is a film set in Central Indiana.}\\
D. \emph{It is a \textbf{forum} focusing on \textbf{green lifestyle}.}\\\hline
\bf Extractive MRC Example \\\hline
\emph{... Early versions were in use by 1851, but the most successful indicator was developed for the high speed engine inventor and manufacturer Charles Porter by Charles Richard and exhibited \textbf{at London Exhibition in 1862}. ...}\\\hline
Q: \emph{\textbf{Where} was the Charles Porter steam engine indicator shown?}\\
Golden Answer: \emph{London Exhibition}\\
Imprecise Answer 1: \emph{London Exhibition \textbf{in 1862}}\\
Imprecise Answer 2: \emph{\textbf{exhibited at} London Exhibition}\\\hline
\end{tabular}
\caption{\label{examples} Different focuses of multi-choice MRC task (RACE) and extractive MRC task (SQuAD 2.0). Texts in bold are the critical information or fallibility parts.}
\end{table}

\begin{figure*}[ht]
\centering
\includegraphics[scale=0.72]{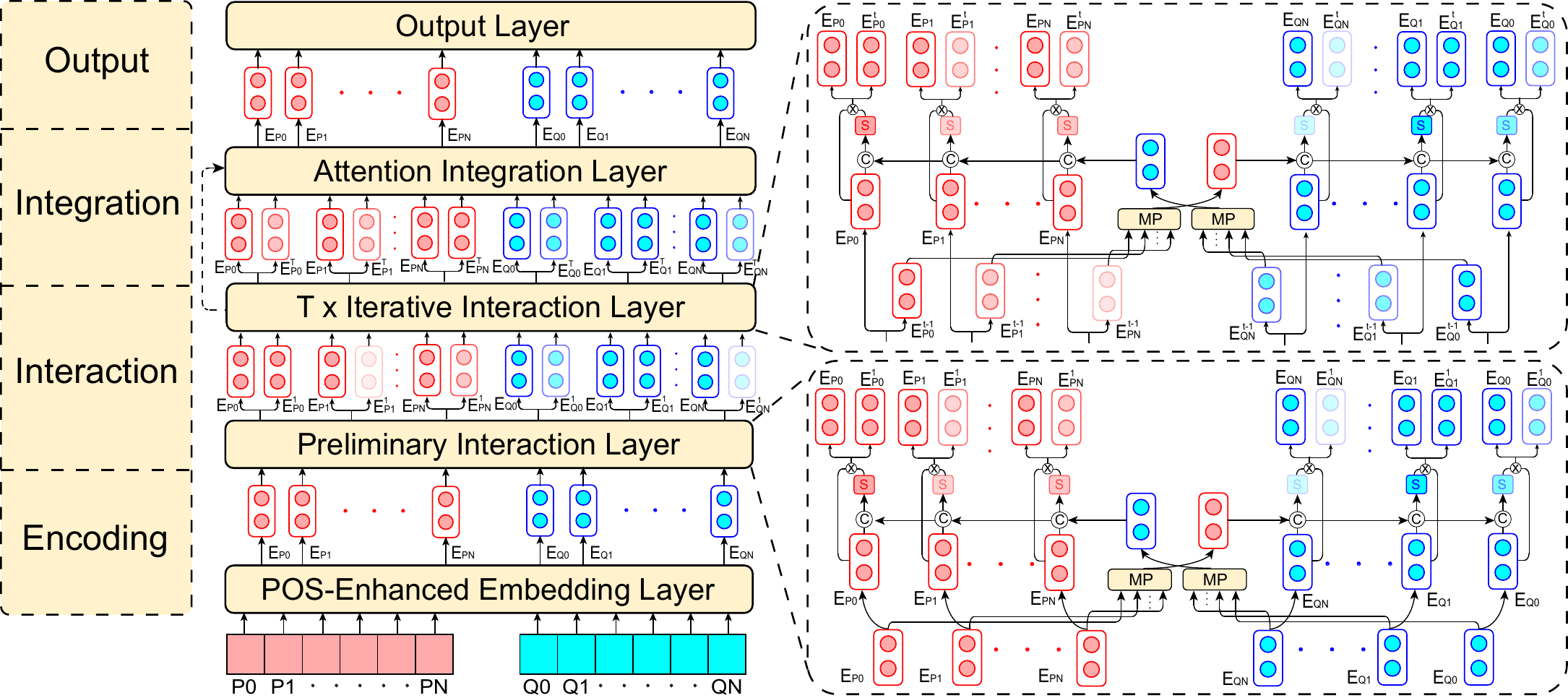}
\caption{Overview of \emph{POI-Net}. $s, c, \times, MP$ donate the normalized attention score, similarity calculation, scalar multiplication, and max pooling operation respectively. The shade of color represent the contribution of corresponding embedding to operating question.}
\label{model}
\end{figure*}

In MRC field, several previous works perform general-purpose language modeling with considerable computing cost at encoding aspect \cite{devlin2019bert, clark2019electra, zhang2020sgnet}, or splice texts among diverse MRC tasks simply to expand training dataset \cite{khashabi2020unifiedqa}, without delicate and specialized design for subcategories in discriminative MRC.
Others utilize excessively detailed design for one special MRC subcategory at decoding aspect \cite{sun2019strategy, zhang2020dcmn}, lacking the universality for overall discriminative MRC.

To fill in above gap in unified modeling for different discriminative MRCs, based on core focuses of extractive and multi-choice MRC, we design two complementary reading strategies at both encoding and decoding aspects.
The encoding design enhances token \textit{linguistic} representation at local level, which is especially effective for extractive MRC.
The explicit possession of word part-of-speech (POS) attribute of human leads to precise answer extraction.
In the extractive sample from Table \ref{examples}, human extracts golden answer span precisely because ``\emph{London Exhibition}” is a proper noun (NNP) corresponding to interrogative qualifier (WDT) ``\textit{Where}” in the question, while imprecise words like ``\textit{1862}” (cardinal number, CD) and ``\textit{exhibited}” (past tense verb, VBD) predicted by machines will not be retained.
Thus, we inject word POS attribute explicitly in embedding form.

The decoding design simulates human \textit{reconsideration} and \textit{integration} abilities at global level, with especial effect for multi-choice MRC.
To handle compound questions with limited attention, human will highlight critical information in turns, and update recognition and attention distribution iteratively.
Inspired by above \textit{reconsideration} strategy, we design \textit{Iterative Co-Attention Mechanism} with no additional parameter, which iteratively executes the interaction between passage and question-option ($Q-O$) pair globally in turns.
In the multi-choice example from Table \ref{examples}, during the first interaction, model may only focus on texts related to rough impression of $Q-O$ pair such as ``\textit{Green Scenes}”, ignoring plentiful but scattered critical information.
But with sufficient iterative interaction, model can ultimately collect all detailed evidence (bold in Table \ref{examples}).
Furthermore, we explore a series of attention \textit{integration} strategies for captured evidence among interaction turns.

We combine two above methods and propose a novel model called \emph{POI-Net} (\textbf{PO}S-Enhanced \textbf{I}terative Co-Attention \textbf{Net}work), to alleviate the gap between machines and humans on discriminative MRC.
We evaluate our model on two multi-choice MRC benchmarks, RACE \cite{lai2017race} and DREAM \cite{sun2019dream}; and two extractive MRC benchmarks, SQuAD 1.1 \cite{rajpurkar2016squad} and SQuAD 2.0 \cite{rajpurkar2018squad2}, obtaining consistent and significant improvements, with nearly zero additional parameters.

\section{Our Model}
We aim to design a lightweight, universal and effective model architecture for various subcategories of discriminative MRC, and the overview of our model is shown in Figure \ref{model}, which consists of four main processes: Encoding (\S \ref{POS-Enhanced Encoder}), Interaction (\S \ref{Iterative Co-Attention Mechanism}), Integration (\S \ref{Attention Integration}) and Output (\S \ref{Adaptation for Discriminative MRC}).

\begin{figure*}[ht]
\centering
\includegraphics[scale=0.87]{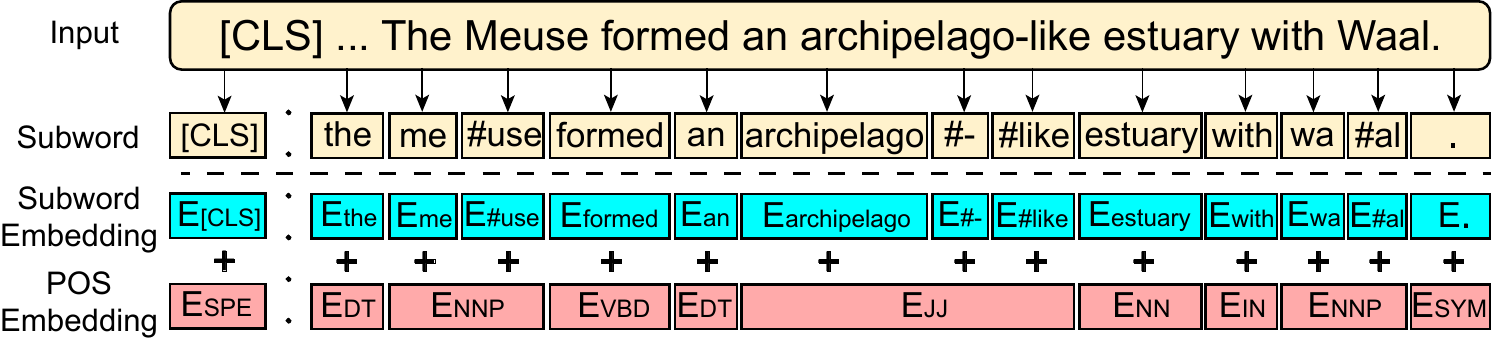}
\caption{The input representation flow of \emph{POI-Net}. The subscripts of \textit{POS Embedding} are POS tags of input words.} 
\label{embedding}
\end{figure*}

\subsection{POS-Enhanced Encoder}
\label{POS-Enhanced Encoder}
Based on pre-trained contextualized encoder ALBERT \cite{lan2019albert}, we encode input tokens with an additional POS embedding layer, as Figure \ref{embedding} shows.
Since the input sequence will be tokenized into subwords in the contextualized encoder, we tokenize sequences in word-level with \textit{nltk} tokenizer \cite{bird2009nltk} additionally and implement \textit{POS-Enhanced Encoder}, where each subword in a complete word will share the same POS tag.

In detail, input sequences are fed into \textit{nltk} POS tagger to obtain the POS tag of each word such as ``JJ”.
Subject to Penn Treebank style, our adopted POS tagger has 36 POS tag types.
Considering on the specific scenarios in discriminative MRC, we add additional $SPE$ tag for special tokens (i.e., $[CLS], [SEP]$), $PAD$ tag for padding tokens and $ERR$ tag for potential unrecognized tokens.
Appendix \ref{Part-Of-Speech Tags List} shows detailed description of POS tags.

The input embedding in our model is the normalized sum of \textit{Subword Embedding} and \textit{POS Embedding}.
Following the basic design in embedding layers of BERT-style models, we retain Token Embedding $E_t$, Segmentation Embedding $E_s$ and Position Embedding $E_p$ in subword-level, constituting \textit{Subword Embedding}.
For \textit{POS Embedding} $E_{POS}$, we implement another embedding layer with the same embedding size to \textit{Subword Embedding}, guaranteeing all above indicator embeddings are in the same vector space.
Formulaically, the input embedding $E$ can be represented as:
$$E=Norm(E_t+E_s+E_p+E_{POS}),$$
where $Norm()$ is a layer normalization function \citep{ba2016norm}.

\subsection{Iterative Co-Attention Mechanism}
\label{Iterative Co-Attention Mechanism}
\emph{POI-Net} employs a lightweight \textit{Iterative Co-Attention} module to simulate human inner reconsidering process, with \textbf{no} additional parameter.

\subsubsection{Preliminary Interaction}
\emph{POI-Net} splits all $N$ input token embeddings into passage domain ($P$) and question (or $Q-O$ pair) domain ($Q$) to start $P-Q$ interactive process.
To generate the overall impression of the given passage or question like humans, \emph{POI-Net} concentrates all embeddings in corresponding domain into one \textit{Concentrated Embedding} by max pooling:
$$CE_P^1=MaxPooling(E_{P0},...,E_{PN})\in \mathbb{R}^H,$$
$$CE_Q^1=MaxPooling(E_{Q0},...,E_{QN})\in \mathbb{R}^H,$$
where $H$ is the hidden size, $PN/QN$ is the token amount of $P/Q$ domain.
Then \emph{POI-Net} calculates the similarity between each token in $E_{P}/E_{Q}$ and $CE_Q^1/CE_P^1$, to generate attention score $s$ for each token contributing to the $P-Q$ pair.
In detail, we use cosine similarly for calculation:
$$s_{P0}^1,...,s_{PN}^1=Cosine([E_{P0},...,E_{PN}],CE_Q^1),$$
$$s_{Q0}^1,...,s_{QN}^1=Cosine([E_{Q0},...,E_{QN}],CE_P^1).$$
We normalize these scores to $[0,1]$ by min-max scaling, then execute dot product with corresponding input embeddings:
$$E_{Pi}^1=\hat s_{Pi}^1 \cdot E_{Pi},\quad E_{Qi}^1=\hat s_{Qi}^1 \cdot E_{Qi},$$
where $\hat s_{Pi}$ is the normalized attention score of $i$-th passage token embedding, $E_{Pi}^1$ is the attention-enhanced embedding of $i$-th passage token after preliminary interaction (the $1$-st turn interaction).

\subsubsection{$t$-th Turn Interaction}
To model human reconsideration ability between passage and question in turns, we add iterable modules with co-attention mechanism, as the \textit{Iterative Interaction Layer} in Figure \ref{model}.
Detailed processes in the $t$-th turn interaction are similar to preliminary interaction:
$$CE_Q^t=MaxPooling(E_{Q0}^{t-1},...,E_{QN}^{t-1})\in \mathbb{R}^H,$$
$$CE_P^t=MaxPooling(E_{P0}^{t-1},...,E_{PN}^{t-1})\in \mathbb{R}^H,$$
$$s_{P0}^t,...,s_{PN}^t=Cosine([E_{P0},...,E_{PN}],CE_Q^t),$$
$$s_{Q0}^t,...,s_{QN}^t=Cosine([E_{Q0},...,E_{QN}],CE_P^t),$$
$$E_{Pi}^t=\hat s_{Pi}^t \cdot E_{Pi},\quad E_{Qi}^t=\hat s_{Qi}^t \cdot E_{Qi}.$$
Note that, during all iteration turns, we calculate attention scores with the original input embedding $E$ instead of attention-enhanced embedding $E^{t-1}$ from the ($t$-1)-th turn, due to:

1) There is no further significant performance improvement by replacing $E$ with $E^{t-1}$ ($<0.2\%$ on base size model), comparing to adopted method;

2) With the same embedding $E$, attention integration in \S \ref{Attention Integration} can be optimized into attention score integration, which is computationally efficient with no additional embedding storage\footnote{Approximate 15.3\% training time is saved on average for each iteration turn.}.

\subsection{Attention Integration}
\label{Attention Integration}
Human recommends to integrate all critical information from multiple turns for a comprehensive conclusion, instead of discarding all findings from previous consideration.
In line with above thought, \emph{POI-Net} returns attention-enhanced embedding $E^t=\hat s^t \cdot E$ of each turn (we only store $\hat s^t$ in an optimized method), and integrates them with specific strategies.
We design four integration strategies according to the contribution proportion of each turn and adopt \textit{Forgetting Strategy} ultimately.

\begin{itemize}
\item \textbf{Average Strategy}: The attention network treats normalized attention score $\hat s^t$ of each turn equally, and produces the ultimate representation vector $\mathbf{R}$ with average value of $\hat s^t$:
$$\mathbf{R}=\frac{1}{T}\sum_{t=1}^T \hat s^t\cdot E\;\in \mathbb{R}^{N\times H},$$
where $T$ is the total amount of iteration turns.

\item \textbf{Weighted Strategy}: The attention network treats $\hat s^t$ with two normalized weighted coefficients $\beta_P^t, \beta_Q^t$, which measure the contribution of the $t$-th turn calculation:
$$\mathbf{R}=\frac{\sum_{t=1}^T \beta_P^t \hat s_P^t}{\sum_{t=1}^T \beta_P^t} \cdot E_P + \frac{\sum_{t=1}^T \beta_Q^t \hat  s_Q^t}{\sum_{t=1}^T \beta_Q^t} \cdot E_Q,$$
$$\tilde \beta_P^t=Max(s_{Q0}^{t-1},...,s_{QN}^{t-1}),$$
$$\tilde \beta_Q^t=Max(s_{P0}^{t-1},...,s_{PN}^{t-1}),$$
$$\beta_P^t = \frac{\tilde \beta_P^t+1}{2},\;\beta_Q^t = \frac{\tilde \beta_Q^t+1}{2},$$
where $s_{Pi}^0=s_{Qi}^0=1.0$.
The design motivation for $\beta_P^t, \beta_Q^t$ is intuitive: when \textit{Concentrated Embedding} $CE_Q^t/CE_P^t$ (calculating attention score at the $t$-th turn) has higher confidence (behaving as higher maximum value in $s_Q^{t-1}/s_P^{t-1}$ due to max pooling calculation), system should pay more attention to input embedding $E_P^t/E_Q^t$ at the $t$-th turn\footnote{Setting $\beta_P^t/\beta_Q^t$ as learnable parameters cannot bring further improvement since the contribution proportion of each turn varies with the specific circumstance of input samples.}.

\item \textbf{Forgetting Strategy}: Since human will partly forget knowledge from previous consideration and focus on findings at current turn, we execute normalization operation of attention scores from two most previous turns iteratively:
$$\mathbf{R}=\frac{\mathbf{s_P^T}+\beta_P^t \hat s_P^T}{1+\beta_P^T} \cdot E_P + \frac{\mathbf{s_Q^T}+\beta_Q^t \hat s_Q^T}{1+\beta_Q^T} \cdot E_Q,$$
$$\mathbf{s_P^T}=\frac{\mathbf{s_P^{T-1}}+\beta_P^t \hat s_P^{T-1}}{1+\beta_P^{T-1}},$$
$$\mathbf{s_Q^T}=\frac{\mathbf{s_Q^{T-1}}+\beta_Q^t \hat s_Q^{T-1}}{1+\beta_Q^{T-1}}.$$
During the iterative normalization, the ultimate proportion of attention scores from previous turns will be diluted gradually, which simulates the effect of forgetting strategy\footnote{Method of activation functions in LSTM \citep{hochreiter1997lstm} may filter out information completely in one single-turn calculation, which cannot bring consistent improvement in our experiments.}.

\item \textbf{Intuition Strategy}: In some cases, human can solve simple questions in intuition without excessive consideration, thus we introduce two attenuation coefficients $\alpha_P^t, \alpha_Q^t$ for attention scores from the $t$-th turn, which decrease gradually as the turn of iteration increases:
$$\mathbf{R}=\frac{\sum_{t=1}^T \alpha_P^t \hat s_P^t}{\sum_{t=1}^T \alpha_P^t} \cdot E_P + \frac{\sum_{t=1}^T \alpha_Q^t \hat s_Q^t}{\sum_{t=1}^T \alpha_Q^t} \cdot E_Q,$$
$$\alpha_P^t=\prod_{i=1}^t\beta_P^i,\;\alpha_Q^t=\prod_{i=1}^t\beta_Q^i.$$
\end{itemize}



\subsection{Adaptation for Discriminative MRC}
\label{Adaptation for Discriminative MRC}

\subsubsection{Multi-choice MRC}
The input sequence for multi-choice MRC is $[CLS]\;P\;[SEP]\;Q+O_i\;[SEP]$, where $+$ denotes concatenation, $O_i$ denotes the $i$-th answer options.
In \textit{Output Layer}, the representation vector $\mathbf{R}\in \mathbb{R}^{N\times H}$ is fed into a max pooling operation to generate general representation:
$$R=MaxPooling(\mathbf{R})\in \mathbb{R}^H.$$
Then a linear softmax layer is employed to calculate probabilities of options, and standard Cross Entropy Loss is employed as the total loss.
Option with the largest probability is determined as the predicted answer.

\subsubsection{Extractive MRC}
The input sequence for extractive MRC can be represented as $[CLS]\;P\;[SEP]\;Q\;[SEP]$, and we use a linear softmax layer to calculate start and end token probabilities in \textit{Output Layer}.
The training object is the sum of Cross Entropy Losses for the start and end token probabilities:
$$\mathcal{L}=y_s\cdot log(s) + y_e\cdot log(e),$$
$$s,e = softmax(Linear(\mathbf{R}))\in \mathbb{R}^N,$$
where $s/e$ are the start/end probabilities for all tokens and $y_s/y_e$ are the start/end targets.

For answer prediction, since some benchmarks have unanswerable questions, we first score the span from the $i$-th token to the $j$-th token as:
$$score_{ij}=s_i+e_j,\quad 0\le i\le j\le N,$$
then the span with the maximum score $score_{has}$ is the predicted answer.
The score of null answer is: $score_{no}=s_0+e_0$, where the $0$-th token is $[CLS]$.
The final score is calculated as $score_{has}-score_{no}$, and a threshold $\delta$ is set to determine whether the question is answerable, which is heuristically computed in linear time.
\emph{POI-Net} predicts the span with the maximum score if the final score is above the threshold, and null answer otherwise.

\section{Experiments}
\subsection{Setup \& Dataset}
The experiments are run on $8$ NVIDIA Tesla P40 GPUs and the implementation of \emph{POI-Net} is based on the Pytorch implementation of ALBERT \cite{paszke2019pytorch}.
We set the maximum iteration turns in \textit{Iterative Co-Attention} as $3$.
Table \ref{hyperparam} shows the hyper-parameters of \emph{POI-Net} achieving reported results.
As a supplement, the warmup rate is 0.1 for all tasks.

\begin{table}[ht]
\small
\fontsize{9pt}{\baselineskip}\selectfont
\centering{
	\begin{tabular}{c|c c c c c}
	    \hline	\bf Hyperparam & LR & MSL & BS & TE & SS\\
		\hline \hline
		\bf DREAM & 1e-5 & 512 & 24 & 4 & 400\\
		\bf RACE & 1e-5 & 512 & 32 & 2 & 4000\\
		\bf SQuAD 1.1 & 1e-5 & 512 & 24 & 2 & 2000\\
		\bf SQuAD 2.0 & 1e-5 & 512 & 24 & 2 & 4000\\\hline
	\end{tabular}
}
\caption{\label{hyperparam}The fine-tuning hyper-parameters of \emph{POI-Net}. LR: learning rate, MSL: maximum sequence length, BS: batch size, TE: training epochs, SS: save steps.}
\end{table}

We evaluate \emph{POI-Net} on two multi-choice MRC benchmarks: RACE \citep{lai2017race}, DREAM \citep{sun2019dream}, and two extractive MRC benchmarks: SQuAD 1.1 \citep{rajpurkar2016squad} and SQuAD 2.0 \citep{rajpurkar2018squad2}.
The detailed introduction is shown as following:

\paragraph{RACE} is a large-scale multi-choice MRC task collected from English examinations which contains nearly 100K questions.
The passages are in the form of articles and most questions need contextual reasoning, and the domains of passages are diversified.

\paragraph{DREAM} is a dialogue-based dataset for multi-choice MRC, containing more than 10K questions.
The challenge of the dataset is that more than 80\% of the questions are non-extractive and require reasoning from multi-turn dialogues.

\paragraph{SQuAD 1.1} is a widely used large-scale extractive MRC benchmark with more than 107K passage-question pairs, which are produced from Wikipedia.
Models are asked to extract precise word span from the Wikipedia passage as the answer of the given passage.

\paragraph{SQuAD 2.0} retains the questions in SQuAD 1.1 with
over 53K unanswerable questions, which are similar to answerable
ones.
For SQuAD 2.0, models must not only answer questions when possible, but also abstain from answering when the question is unanswerable with the paragraph.

\subsection{Results}

\begin{table*} [htbp]
\small
\fontsize{9pt}{\baselineskip}\selectfont
	\centering{
		\begin{tabular}{p{4.5cm}|c c|c c|c c|c c}
			\hline\multirow{2}{*}{\textbf{Model}} & \multicolumn{2}{c|}{\bf DREAM} & \multicolumn{2}{c|}{\bf RACE} & \multicolumn{2}{c|}{\bf SQuAD 1.1} & \multicolumn{2}{c}{\bf SQuAD 2.0}\\
			~ & \bf Dev & \bf Test & \bf Dev(M/H) &\bf Test(M/H) & \bf EM & \bf F1& \bf EM & \bf F1\\\hline\hline
			BERT$_{base}$ \cite{devlin2019bert}& 63.4 & 63.2 & 64.6 (-- / --) & 65.0 (71.1 / 62.3) & 80.8 & 88.5 & 77.6 & 80.4\\
			ALBERT$_{base}$ \cite{lan2019albert} & 64.5 & 64.4 & 64.0 (-- / --) & -- (-- / --) & 82.3 & 89.3 & 77.1 & 80.0\\
			BERT$_{large}$ \cite{devlin2019bert} & 66.0 & 66.8 & 72.7 (76.7 / 71.0) & 72.0 (76.6 / 70.1) & 85.5 & 92.2 & 82.2 & 85.0\\
			SG-Net \cite{zhang2020sgnet} & -- & -- & -- (-- / --) & 74.2 (78.8 / 72.2) & -- & -- & 85.6 & 88.3\\
			RoBERTa$_{large}$ \cite{liu2019roberta} & 85.4 & 85.0 & -- (-- / --) & 83.2 (86.5 / 81.8) & -- & -- & 86.5 & 89.4\\
			RoBERTa$_{large}$+MMM \cite{jin2019mmm} & 88.0 & 88.9 & -- (-- / --) & 85.0 (89.1 / 83.3) & -- & -- & -- & --\\
			ALBERT$_{xxlarge}$ \cite{lan2019albert} & 89.2 & 88.5 & -- (-- / --) & 86.5 (89.0 / 85.5) & 88.3 & 94.1 & 85.1 & 88.1\\
			ALBERT$_{xxlarge}$ + DUMA \cite{zhu2020dual} & 89.9 & 90.4  & 88.1 (-- / --) & 88.0 (90.9 / 86.7) & -- & -- & -- & --\\\hline
			ALBERT$_{base}$ (rerun) & 65.7 & 65.6 & 67.9 (72.3 / 65.7) & 67.2 (72.1 / 65.2) & 82.7 & 89.9 & 77.9 & 81.0\\
			POI-Net on ALBERT$_{base}$ & 68.6 & 68.5 & 72.4 (76.3 / 70.0) & 71.0 (75.7 / 69.0) & 84.5 & 91.3 & 79.5 & 82.7\\\hline
			ALBERT$_{xxlarge}$ (rerun) & 88.7 & 88.3 & 86.6 (89.4 / 85.2) & 86.5 (89.2 / 85.4) & 88.2 & 93.9 & 85.4 & 88.5\\
			POI-Net on ALBERT$_{xxlarge}$ & \bf 90.0 & 90.3 &\bf 88.1 (91.2 / 86.3) &\bf 88.3 (91.5 / 86.8) &\bf 89.5 &\bf 95.0 &\bf 87.7 &\bf 90.6\\\hline
		\end{tabular}
	}
	\caption{\label{result} Results of BERT-style models on DREAM, RACE, SQuAD 1.1 and SQuAD 2.0. Results in the first domain are from the leaderboards and corresponding papers\footnote{Due to the test sets of SQuAD 1.1 and SQuAD 2.0 are not open for free evaluation with different random seeds, we report the results on development set instead.}.}
\end{table*}

We take accuracy as evaluation criteria for multi-choice benchmarks, while exact match (EM) and a softer metric F1 score for extractive benchmarks.
The average results of three random seeds are shown in Table \ref{result}, where we only display several BERT-style models with comparable parameters.
Appendix \ref{Complete Comparison Results on Benchmarks} reports the complete comparison results with other public works on each benchmark.

The results show that, for multi-choice benchmarks, our model outperforms most baselines and comparison works, and passes the significance test \cite{zhang2020retro} with $p-value < 0.01$ in DREAM (2.0\% average improvement) and RACE (1.7\% average improvement).
And for extractive benchmarks, though the performance of baseline ALBERT is strong, our model still boosts it essentially (1.3\% average improvement on EM for SQuAD 1.1 and 2.3\% for SQuAD 2.0).
Furthermore, we report the parameter scale and training/inference time costs in \S \ref{Role of Iterative Co-Attention}.

\section{Ablation Studies}
In this section, we implement \emph{POI-Net} on ALBERT$_{base}$ for further discussions, and such settings have the similar quantitative tendency to \emph{POI-Net} on ALBERT$_{xxlarge}$.

\subsection{Ablation}
\begin{table}[ht]
\small
\fontsize{9pt}{\baselineskip}\selectfont
	\centering{
		\begin{tabular}{l|c|c c}
		\hline\multirow{2}{*}{\textbf{Model}} & \textbf{RACE} & \multicolumn{2}{c}{\bf SQuAD 1.1}\\
		~ & \bf Acc & \bf EM & \bf F1\\\hline\hline
        Baseline (ALBERT$_{base}$) & 67.88 & 82.66 & 89.91\\
        POI-Net on ALBERT$_{base}$ & 72.44 & 84.48 & 91.28\\
        \quad - POS Embedding & 71.74 & 83.51 & 90.64\\
        \quad - Iterative Co-Attention & 69.02 & 83.65 & 90.77\\\hline
        Baseline (rerun BERT$_{base}$) & 64.73 & 81.21 & 88.84\\
        POI-Net on BERT$_{base}$ & 68.02 & 83.43 & 90.47\\\hline
		\end{tabular}
	}
	\caption{\label{ablation_result} Ablation studies on RACE and SQuAD 1.1.}
\end{table}

To evaluate the contribution of each component in \emph{POI-Net}, we perform ablation studies on RACE and SQuAD 1.1 development sets and report the average results of three random seeds in Table \ref{ablation_result}.
The results indicate that, both \textit{POS Embedding} and \textit{Iterative Co-Attention Mechanism} provide considerable contributions to \emph{POI-Net}, but in different roles for certain MRC subcategory.

For multi-choice MRC like RACE, \textit{Iterative Co-Attention Mechanism} contributes much more than \textit{POS Embedding} (3.86\% v.s. 1.14\%), since multi-choice MRC requires to highlight and integrate critical information in passages comprehensively.
Therefore, potential omission of critical evidence may be fatal for answer prediction, which is guaranteed by \textit{Iterative Co-Attention Mechanism}, while precise evidence span boundary and POS attributes are not as important as the former.

On the contrary, simple \textit{POS Embedding} even brings a little more improvement than the well-designed \textit{Iterative Co-Attention} (0.99\% v.s. 0.85\% on EM) for extractive MRC.
In these tasks, model focuses on answer span extraction with precise boundaries, and requires to discard interference words which not exactly match questions, such as redundant verbs, prepositions and infinitives (``\textit{politically and socially unstable}” instead of ``\textit{\textbf{to be} politically and socially unstable}”), or partial interception of proper nouns (``\textit{Seljuk Turks}” instead of ``\textit{Turks}”).
With the POS attribute of each word, \emph{POI-Net} locates the boundaries of answer spans precisely\footnote{Note that, the improvement of \emph{POI-Net} on EM score is consistently higher than F1 score, as corroboration.}.
Since extractive MRC does not require comprehensive information integration like multi-choice MRC, the improvement from \textit{Iterative Co-Attention Mechanism} is less significant.

Besides, we also implement \emph{POI-Net} on other contextualized encoders like BERT, and achieve significant improvements as Table \ref{ablation_result} shows.
The consistent and significant improvements over various baselines verify the universal effectiveness of \emph{POI-Net}.

\subsection{Role of POS Embedding}
\begin{table}[ht]
\small
\fontsize{9pt}{\baselineskip}\selectfont
	\centering{
		\begin{tabular}{c|r|r|r}
		\hline\textbf{POS Type} & \textbf{Golden Answer} & \textbf{POI-Net} & \textbf{Baseline}\\\hline\hline
        NN & 11192 & 11254 & 11504\\
        CD & 3511 & 3723 & 3816\\
        NNS & 2875 & 2812 & 2743\\
        JJ & 1654 & 1671 & 1774\\
        IN & 396 & 308 & 242\\
        VBN & 348 & 321 & 299\\
        RB & 339 & 315 & 284\\
        VBG & 331 & 328 & 293\\\hline
		\end{tabular}
	}
	\caption{\label{pos_tags} The POS type statistics of boundary words in golden answer, predicted answer by \emph{POI-Net} and baseline ALBERT$_{base}$. We only display POS types whose occurrence is higher than 300.}
\end{table}

\begin{figure}[ht]
\centering
\includegraphics[scale=0.52]{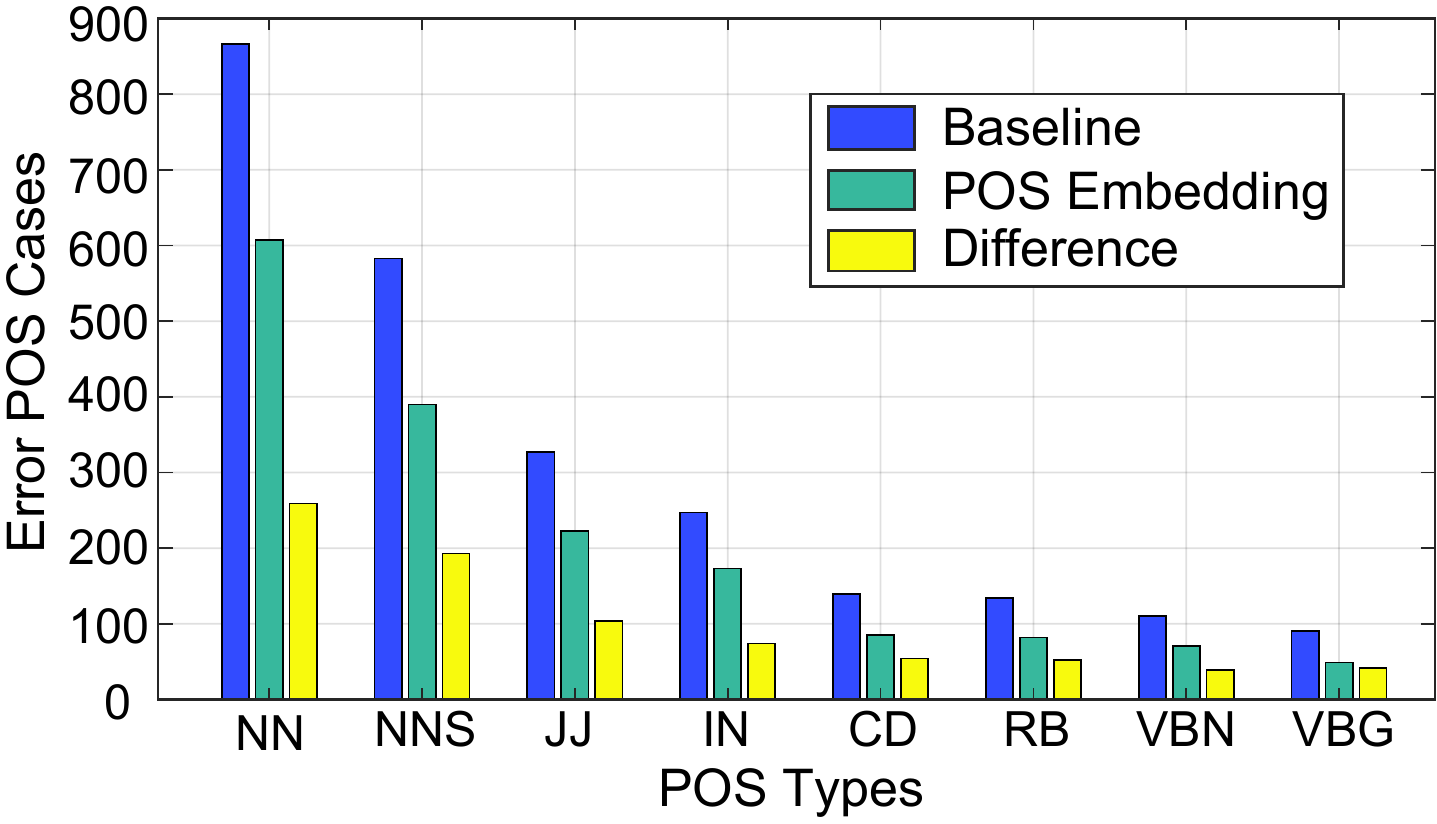}
\caption{Error POS classification case statistics of \emph{POI-Net} and baseline. For explanation, the first square pillar (Height: 866) means, there are 866 cases whose POS type of boundary word in golden answer is ``NN”, but the baseline predicts an error word in a non-``NN” type.}
\label{pos_diff}
\end{figure}

To study how \textit{POS Embedding} enhances token representation, we make a series of statistics on SQuAD 1.1 development set about:
1) POS type of boundary words from predicted spans, as Table \ref{pos_tags} shows;
2) error POS classification of \emph{POI-Net} and its baseline ALBERT$_{base}$, as Figure \ref{pos_diff} shows.

The statistical results show, with \textit{POS Embedding}, the overall distribution of the POS types of answer boundary words predicted by \emph{POI-Net} is more similar to golden answer, compared with its baseline;
and the amount of error POS classification cases by \emph{POI-Net} also reduces significantly.
And there are also two further findings:

1) The correction proportion of error POS classification (8.09\%) is much higher than correction proportion of overall error predictions (1.82\%) in \emph{POI-Net}, which indicates the correction of POS classification benefits mostly from the perception of word POS attributes by \textit{POS Embedding}, instead of the improvement on overall accuracy.

2) Though answers in SQuAD 1.1 incline to distribute in several specific POS types (``NN”, ``CD”, ``NNS” and ``JJ”), \textit{POS Embedding} prompts model to consider words in each POS type more equally than the baseline, and the predicted proportions of words in rarer POS type (``IN”, ``VBN”, ``RB”, ``VBG” and so on) increase.

\subsection{Research on the Robustness of POS Embedding}
Robustness is one of the important indicators to measure model performance, when there is numerous rough data or resource in applied tasks.
To measure the anti-interference of \textit{POS Embedding}, we randomly modify part of POS tags from \textit{nltk} POS tagger to error tags, and the results on SQuAD 1.1 development set are shown in Table \ref{robustness}.

\begin{table}[ht]
\small
\fontsize{9pt}{\baselineskip}\selectfont
	\centering{
		\begin{tabular}{p{4cm}|c c}
		\hline\textbf{Model} & \bf EM & \bf F1\\\hline\hline
        Baseline (ALBERT$_{base}$) & 82.66 & 89.91\\
        POI-Net on ALBERT$_{base}$ & 84.48 & 91.28\\\hline
        \quad 5\% error POS tags & 84.35 & 91.21\\
        \quad 10\% error POS tags & 84.06 & 91.05\\
        \quad 20\% error POS tags & 83.87 & 90.80\\
        \quad - POS Embedding & 83.51 & 90.64\\\hline
		\end{tabular}
	}
	\caption{\label{robustness} Results of robustness research of POS Embedding on dev sets from SQuAD 1.1.}
\end{table}

The results indicate that, \textit{POI-Net} possesses satisfactory \textit{POS Embedding} robustness, and the improvement brought by \textit{POS Embedding} will not suffer a lot with a slight disturbance (5\%).
We argue that the robustness of \textit{POI-Net} may benefit from the integration with other contextualized embeddings, such as Token Embedding $E_t$ which encodes the contextual meaning of current word or subword.
Though more violent interference (20\%) may further hurt token representations, existing mature POS taggers achieve 97\% + accuracy, which can prevent the occurrence of above situations.

\subsection{Role of Iterative Co-Attention Mechanism}
\label{Role of Iterative Co-Attention}
\begin{figure}[ht]
\centering
\includegraphics[scale=0.45]{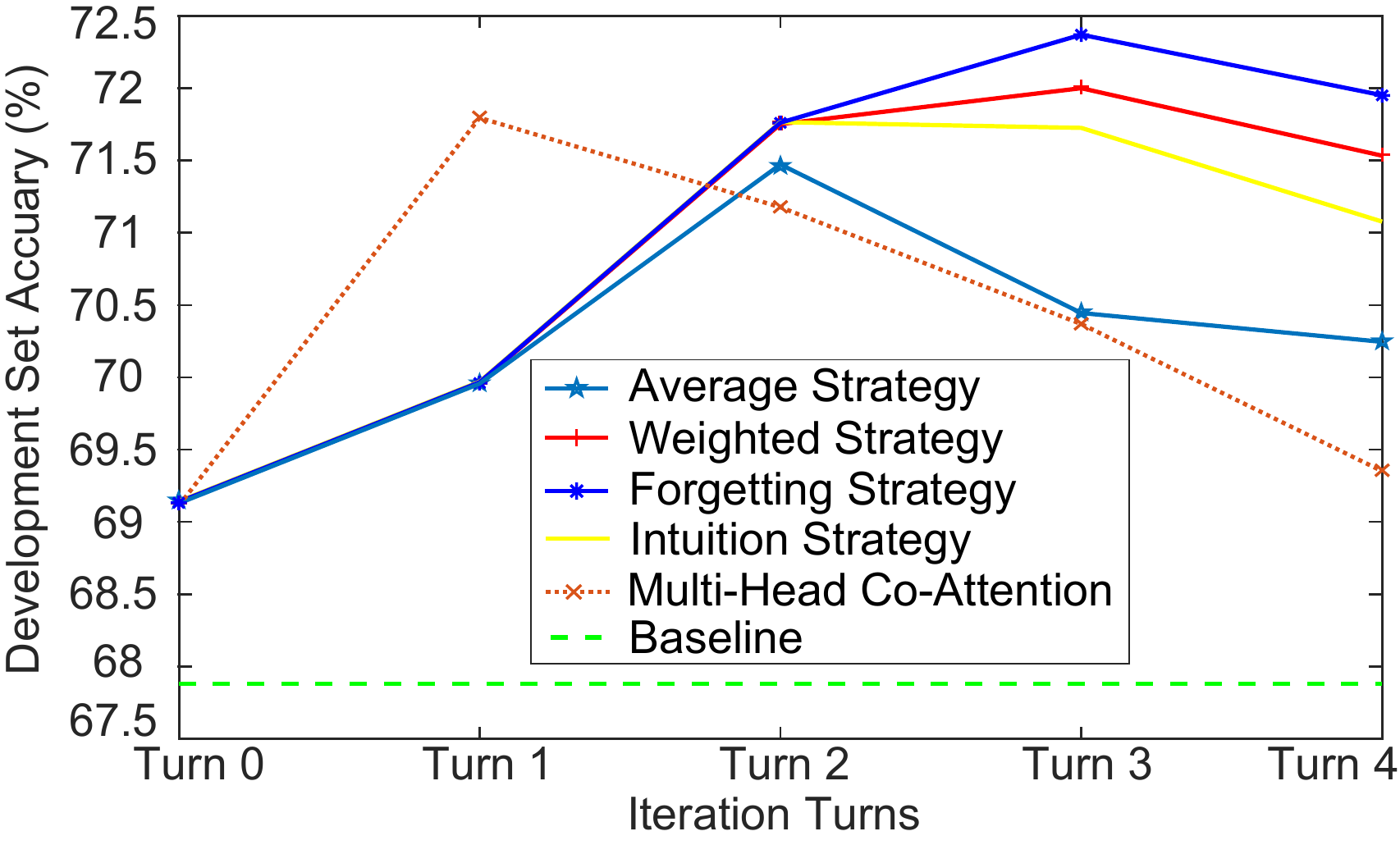}
\caption{Comparative experiments on \textit{Iterative Co-Attention Mechanism}. When iteration turn is 0, the model is equivalent to baseline with \textit{POS Embedding}.}
\label{strategy}
\end{figure}

To explore the most suitable integration strategy and maximum iteration turn in \textit{Iterative Co-Attention Mechanism}, we implement our proposed strategies with different maximum iteration turns, together with a baseline replacing \textit{Iterative Co-Attention} mechanism by a widely used Multi-head Co-Attention mechanism \cite{devlin2019bert, zhang2020dcmn, zhang2020retro} for comparison in Figure \ref{strategy}.
We take RACE as the evaluated benchmark due to the significant effect of attention mechanism to multi-choice MRC.

As the figure shows, forgetting strategy leads to the best performance, with slight improvement than weighted strategy.
Both these two strategies are in line with the logical evidence integration in human reconsidering process, while average strategy and intuition strategy may work against common human logic.
From the trends of four strategies in multiple iterations, we conclude that 2 or 3 iteration turns for \textit{Iterative Co-Attention} lead to an appropriate result, due to:

1) Fewer iteration turns may lead to inadequate interaction between passage and question, and model may focus on rough cognition instead of exhaustive critical information;

2) Excessive iteration turns may lead to over-integration of information, declining the contribution by real critical evidence.

Compared to the typical Multi-head Co-Attention mechanism, our proposed \textit{Iterative Co-Attention} mechanism obtains higher performance with more iterations, indicating it has stronger iterative reconsideration ability.

\begin{table}[ht]
\small
\fontsize{9pt}{\baselineskip}\selectfont
	\centering{
		\begin{tabular}{p{4.3cm}|c}
		    \hline	\bf Model & \bf Parameters\\
			\hline \hline
            ALBERT$_{base}$ \citep{lan2019albert}& 12M\\
            ALBERT$_{base}$ (rerun) & 11.14M\\
            Multi-head Co-Attention on ALBERT$_{base}$ & 17.94M\\
            POI-Net on ALBERT$_{base}$ & \bf 11.15M\\\hline
            ALBERT$_{xxlarge}$ \citep{lan2019albert}& 235M\\
            ALBERT$_{xxlarge}$ (rerun) & 212.29M\\
            Multi-head Co-Attention on ALBERT$_{xxlarge}$ & 404.50M\\
            POI-Net on ALBERT$_{xxlarge}$ & \bf 212.30M\\\hline
		\end{tabular}
	}
	\caption{\label{parameters} Training parameters in \emph{POI-Net} and baselines.}
\end{table}

Besides, \textit{Iterative Co-Attention} defeats Multi-head Co-Attention on both parameter size and training time cost.
As the parameter comparison in Table \ref{parameters} shows, \emph{POI-Net} basically brings no additional parameter except an linear embedding layer for \textit{POS Embedding}.
Multi-head Co-Attention mechanism and models based on it (like DUMA in Table \ref{result}) introduces much more parameters, with slightly lower performance.
We also record time costs on RACE for one training epoch on ALBERT$_{base}$, \textit{Iterative Co-Attention} costs $54,62,72,83,96$ minutes from $0$-turn iteration to $4$-turn iterations, while Multi-head Co-Attention costs $54,65,76,89,109$ minutes instead, with $8.3\%$ increase on average.


\subsection{Visualization}
\begin{figure*}[ht]
\centering
\includegraphics[scale=0.8]{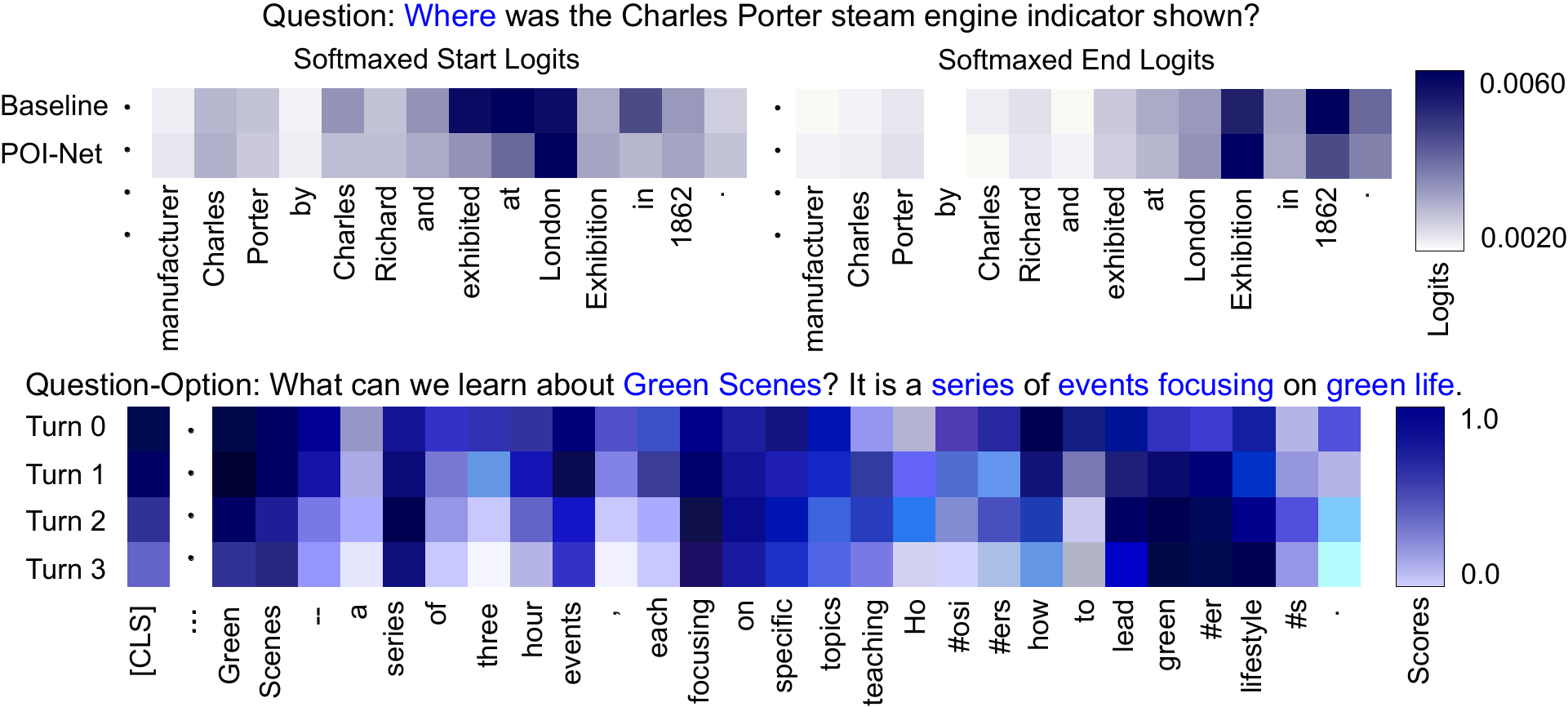}
\caption{Visualization of \emph{POI-Net} and its baseline on extractive example (upper) and multi-choice example (lower) in Table \ref{examples}. The indicator for extractive example is softmaxed logit, and for multi-choice example is normalized attention score $\hat s_P^t$.}
\label{visual}
\vspace{-0.3cm}
\end{figure*}

We perform a visualization display for discriminative MRC examples in Table \ref{examples}, as Figure \ref{visual} shows.
For the extractive example, benefited from \textit{POS Embedding}, \emph{POI-Net} predicts the precise answer span, based on the interrogative qualifier ``\textit{where}” and POS attributes of controversial boundary tokens \textit{``exhibited”, ``at”, ``London”, ``Exhibition”, ``1862”}.

And for the multi-choice example, without proposed \textit{Iterative Co-Attention Mechanism}, the overall distribution of attention is more scattered.
The baseline can only notice special tokens like $[CLS]$ at the $0$-th turn, and even interrogative qualifier ``\textit{how}” due to the similar usage to ``\textit{what}” in the question.
With the execution of \textit{Iterative Co-Attention}, \emph{POI-Net} pays more attention on discrete critical words like ``\textit{Green Scenes}” and ``\textit{events}” at the $1$-st turn, ``\textit{series}” and ``\textit{focusing}” at the $2$-nd turn and ``\textit{greener lifestyle}” at the $3$-rd turn.
After the integration of all above critical evidence, \emph{POI-Net} predicts the golden option ultimately.

\section{Related Studies}
\subsection{Semantic and Linguistic Embedding}
To cope with challenging MRC tasks, numerous powerful pre-trained language models (PLMs) have been proposed \cite{devlin2019bert, lewis2020bart, raffel2020t5}.
Though advanced PLMs demonstrate strong ability in contextual representation, the lack of explicit \textit{semantic} and \textit{linguistic} clues leads to the bottleneck of previous works.

Benefited from the development of semantic role labeling \cite{li2018srl} and dependency syntactic parsing \cite{zhou2019head}, some researchers focus on enhancing semantic representations.
\citet{zhang2020semBERT} strengthen token representation by fusing semantic role labels, while \citet{zhang2020sgnet} and \citet{bai2021syntax} implement additional self attention layers to encode syntactic dependency.
Furthermore, \citet{mihaylov2019discourse} employ multiple discourse-aware semantic annotations for MRC on narrative texts. 

Instead of semantic information, we pay attention to more accessible part-of-speech (POS) information, which has been widely used into non-MRC fields, such as open domain QA \cite{chen2017odqa}, with much lower pre-processing calculation consumption but higher accuracy \cite{Bohnet2018pos, Strubell2018linguistically, zhou2020limit}.
However, previous application of POS attributes mostly stays in primitive and rough embedding methods \cite{huang2018fusionnet}, leading to much slighter improvement than proposed \emph{POI-Net}.

\subsection{Attention Mechanism}
In discriminative MRC field, various attention mechanisms \cite{Raffel2015self,seo2017bidaf,wang2017gated,vaswani2017attention} play increasingly important roles.
Initially, attention mechanism is mainly adopted on extractive MRC \cite{wei2018qanet, cui2021attmrc}, such as multiple polishing of answer spans \cite{xiong2017coattention} and multi-granularity representations generation \cite{zheng2020document,chen2020biatt}.
Recently, researchers notice its special effect for multi-choice MRC.
\citet{zhang2020dcmn} model domains bidirectionally with dual co-matching network, \citet{jin2019mmm} use multi-step attention as classifier, and \citet{zhu2020dual} design multi-head co-attentions for collaborative interactions.

We thus propose a universal \textit{Iterative Co-Attention} mechanism, which performs interaction between paired input domains iteratively, to hopefully enhance discriminative MRC.
Unlike other works introducing numerous parameters by complicated attention network \cite{zhang2020dcmn}, our \textit{POI-Net} is more effective and efficient with almost no introduction of additional parameters.

\section{Conclusion}
In this work, we propose \textbf{PO}S-Enhanced \textbf{I}terative Co-Attention \textbf{Net}work (\textit{POI-Net}), as a lightweight unified modeling for multiple subcategories of discriminative MRC.
\textit{POI-Net} utilizes \textit{POS Embedding} to encode POS attributes for the preciseness of answer boundary, and \textit{Iterative Co-Attention Mechanism} with integration strategy is employed to highlight and integrate critical information at decoding aspect, with almost no additional parameter.
As the first effective and unified modeling with pertinence for different types of discriminative MRC, evaluation results on four extractive and multi-choice MRC benchmarks consistently indicate the general effectiveness and applicability of our model.

\bibliography{anthology,custom}

\begin{thebibliography}{55}
\expandafter\ifx\csname natexlab\endcsname\relax\def\natexlab#1{#1}\fi

\bibitem[{Ba et~al.(2016)Ba, Kiros, and Hinton}]{ba2016norm}
Jimmy~Lei Ba, Jamie~Ryan Kiros, and Geoffrey~E. Hinton. 2016.
\newblock \href {http://arxiv.org/abs/1607.06450} {Layer normalization}.

\bibitem[{Back et~al.(2020)Back, Chinthakindi, Kedia, Lee, and
  Choo}]{back2020NeurQuRI}
Seohyun Back, Sai~Chetan Chinthakindi, Akhil Kedia, Haejun Lee, and Jaegul
  Choo. 2020.
\newblock Neurquri: Neural question requirement inspector for answerability
  prediction.
\newblock In \emph{ICLR}.

\bibitem[{Bai et~al.(2021)Bai, Wang, Chen, Yang, Bai, Yu, and
  Tong}]{bai2021syntax}
Jiangang Bai, Yujing Wang, Yiren Chen, Yaming Yang, Jing Bai, Jing Yu, and
  Yunhai Tong. 2021.
\newblock \href {https://aclanthology.org/2021.eacl-main.262} {Syntax-{BERT}:
  Improving pre-trained transformers with syntax trees}.
\newblock In \emph{Proceedings of the 16th Conference of the European Chapter
  of the Association for Computational Linguistics: Main Volume}, pages
  3011--3020, Online. Association for Computational Linguistics.

\bibitem[{Baradaran et~al.(2020)Baradaran, Ghiasi, and
  Amirkhani}]{baradaran2020survey}
Razieh Baradaran, Razieh Ghiasi, and Hossein Amirkhani. 2020.
\newblock A survey on machine reading comprehension systems.
\newblock \emph{arXiv preprint arXiv:2001.01582}.

\bibitem[{Bird et~al.(2009)Bird, Klein, and Loper}]{bird2009nltk}
Steven Bird, Ewan Klein, and Edward Loper. 2009.
\newblock \emph{Natural language processing with Python: analyzing text with
  the natural language toolkit}.
\newblock " O'Reilly Media, Inc.".

\bibitem[{Bohnet et~al.(2018)Bohnet, McDonald, Sim{\~o}es, Andor, Pitler, and
  Maynez}]{Bohnet2018pos}
Bernd Bohnet, Ryan McDonald, Gon{\c{c}}alo Sim{\~o}es, Daniel Andor, Emily
  Pitler, and Joshua Maynez. 2018.
\newblock \href {https://doi.org/10.18653/v1/P18-1246} {Morphosyntactic tagging
  with a meta-{B}i{LSTM} model over context sensitive token encodings}.
\newblock In \emph{Proceedings of the 56th Annual Meeting of the Association
  for Computational Linguistics (Volume 1: Long Papers)}, pages 2642--2652,
  Melbourne, Australia. Association for Computational Linguistics.

\bibitem[{Chen et~al.(2017)Chen, Fisch, Weston, and Bordes}]{chen2017odqa}
Danqi Chen, Adam Fisch, Jason Weston, and Antoine Bordes. 2017.
\newblock \href {https://doi.org/10.18653/v1/P17-1171} {Reading {W}ikipedia to
  answer open-domain questions}.
\newblock In \emph{Proceedings of the 55th Annual Meeting of the Association
  for Computational Linguistics (Volume 1: Long Papers)}, pages 1870--1879,
  Vancouver, Canada. Association for Computational Linguistics.

\bibitem[{Chen et~al.(2020)Chen, Liu, You, Zhou, and Zou}]{chen2020biatt}
Nuo Chen, Fenglin Liu, Chenyu You, Peilin Zhou, and Yuexian Zou. 2020.
\newblock \href {http://arxiv.org/abs/2012.10877} {Adaptive bi-directional
  attention: Exploring multi-granularity representations for machine reading
  comprehension}.

\bibitem[{Clark et~al.(2020)Clark, Luong, Le, and Manning}]{clark2019electra}
Kevin Clark, Minh{-}Thang Luong, Quoc~V. Le, and Christopher~D. Manning. 2020.
\newblock \href {https://openreview.net/forum?id=r1xMH1BtvB} {{ELECTRA:}
  pre-training text encoders as discriminators rather than generators}.
\newblock In \emph{8th International Conference on Learning Representations,
  {ICLR} 2020, Addis Ababa, Ethiopia, April 26-30, 2020}. OpenReview.net.

\bibitem[{Cui et~al.(2021)Cui, Zhang, Che, Liu, and Chen}]{cui2021attmrc}
Yiming Cui, Wei-Nan Zhang, Wanxiang Che, Ting Liu, and Zhigang Chen. 2021.
\newblock \href {http://arxiv.org/abs/2108.11574} {Understanding attention in
  machine reading comprehension}.

\bibitem[{Devlin et~al.(2019)Devlin, Chang, Lee, and
  Toutanova}]{devlin2019bert}
Jacob Devlin, Ming-Wei Chang, Kenton Lee, and Kristina Toutanova. 2019.
\newblock \href {https://doi.org/10.18653/v1/N19-1423} {{BERT}: Pre-training of
  deep bidirectional transformers for language understanding}.
\newblock In \emph{Proceedings of the 2019 Conference of the North {A}merican
  Chapter of the Association for Computational Linguistics: Human Language
  Technologies, Volume 1 (Long and Short Papers)}, pages 4171--4186,
  Minneapolis, Minnesota. Association for Computational Linguistics.

\bibitem[{Hochreiter and Schmidhuber(1997)}]{hochreiter1997lstm}
S.~Hochreiter and J.~Schmidhuber. 1997.
\newblock Long short-term memory.
\newblock \emph{Neural Computation}, 9(8):1735--1780.

\bibitem[{Hu et~al.(2018)Hu, Peng, Huang, Qiu, Wei, and Zhou}]{hu2018rm}
Minghao Hu, Yuxing Peng, Zhen Huang, Xipeng Qiu, Furu Wei, and Ming Zhou. 2018.
\newblock Reinforced mnemonic reader for machine reading comprehension.
\newblock In \emph{IJCAI}.

\bibitem[{Huang et~al.(2018)Huang, Zhu, Shen, and Chen}]{huang2018fusionnet}
Hsin-Yuan Huang, Chenguang Zhu, Yelong Shen, and Weizhu Chen. 2018.
\newblock \href {https://openreview.net/forum?id=BJIgi_eCZ} {Fusionnet: Fusing
  via fully-aware attention with application to machine comprehension}.
\newblock In \emph{International Conference on Learning Representations}.

\bibitem[{Huang et~al.(2019)Huang, Le~Bras, Bhagavatula, and
  Choi}]{huang2019CosmosQA}
Lifu Huang, Ronan Le~Bras, Chandra Bhagavatula, and Yejin Choi. 2019.
\newblock \href {https://doi.org/10.18653/v1/D19-1243} {Cosmos {QA}: Machine
  reading comprehension with contextual commonsense reasoning}.
\newblock In \emph{Proceedings of the 2019 Conference on Empirical Methods in
  Natural Language Processing and the 9th International Joint Conference on
  Natural Language Processing (EMNLP-IJCNLP)}, pages 2391--2401, Hong Kong,
  China. Association for Computational Linguistics.

\bibitem[{Jiang et~al.(2020)Jiang, Wu, Gong, Cheng, Meng, Lin, Chen, and
  li}]{jiang2020single}
Yufan Jiang, Shuangzhi Wu, Jing Gong, Yahui Cheng, Peng Meng, Weiliang Lin,
  Zhibo Chen, and Mu~li. 2020.
\newblock \href {http://arxiv.org/abs/2011.03292} {Improving machine reading
  comprehension with single-choice decision and transfer learning}.

\bibitem[{Jin et~al.(2020)Jin, Gao, Kao, Chung, and Hakkani-tur}]{jin2019mmm}
Di~Jin, Shuyang Gao, Jiun-Yu Kao, Tagyoung Chung, and Dilek Hakkani-tur. 2020.
\newblock Mmm: Multi-stage multi-task learning for multi-choice reading
  comprehension.
\newblock In \emph{AAAI}.

\bibitem[{Joshi et~al.(2020)Joshi, Chen, Liu, Weld, Zettlemoyer, and
  Levy}]{joshi2020spanbert}
Mandar Joshi, Danqi Chen, Yinhan Liu, Daniel~S. Weld, Luke Zettlemoyer, and
  Omer Levy. 2020.
\newblock \href {https://doi.org/10.1162/tacl_a_00300} {{S}pan{BERT}: Improving
  pre-training by representing and predicting spans}.
\newblock \emph{Transactions of the Association for Computational Linguistics},
  8:64--77.

\bibitem[{Joshi et~al.(2017)Joshi, Choi, Weld, and
  Zettlemoyer}]{joshi2017triviaqa}
Mandar Joshi, Eunsol Choi, Daniel Weld, and Luke Zettlemoyer. 2017.
\newblock \href {https://doi.org/10.18653/v1/P17-1147} {{T}rivia{QA}: A large
  scale distantly supervised challenge dataset for reading comprehension}.
\newblock In \emph{Proceedings of the 55th Annual Meeting of the Association
  for Computational Linguistics (Volume 1: Long Papers)}, pages 1601--1611,
  Vancouver, Canada. Association for Computational Linguistics.

\bibitem[{Khashabi et~al.(2018)Khashabi, Chaturvedi, Roth, Upadhyay, and
  Roth}]{Khashabi2018MultiRC}
Daniel Khashabi, Snigdha Chaturvedi, Michael Roth, Shyam Upadhyay, and Dan
  Roth. 2018.
\newblock \href {https://doi.org/10.18653/v1/N18-1023} {Looking beyond the
  surface: A challenge set for reading comprehension over multiple sentences}.
\newblock In \emph{Proceedings of the 2018 Conference of the North {A}merican
  Chapter of the Association for Computational Linguistics: Human Language
  Technologies, Volume 1 (Long Papers)}, pages 252--262, New Orleans,
  Louisiana. Association for Computational Linguistics.

\bibitem[{Khashabi et~al.(2020)Khashabi, Min, Khot, Sabharwal, Tafjord, Clark,
  and Hajishirzi}]{khashabi2020unifiedqa}
Daniel Khashabi, Sewon Min, Tushar Khot, Ashish Sabharwal, Oyvind Tafjord,
  Peter Clark, and Hannaneh Hajishirzi. 2020.
\newblock \href {https://doi.org/10.18653/v1/2020.findings-emnlp.171}
  {{UNIFIEDQA}: Crossing format boundaries with a single {QA} system}.
\newblock In \emph{Findings of the Association for Computational Linguistics:
  EMNLP 2020}, pages 1896--1907, Online. Association for Computational
  Linguistics.

\bibitem[{Ko{\v{c}}isk{\'y} et~al.(2018)Ko{\v{c}}isk{\'y}, Schwarz, Blunsom,
  Dyer, Hermann, Melis, and Grefenstette}]{Schwarz2018narrativeqa}
Tom{\'a}{\v{s}} Ko{\v{c}}isk{\'y}, Jonathan Schwarz, Phil Blunsom, Chris Dyer,
  Karl~Moritz Hermann, G{\'a}bor Melis, and Edward Grefenstette. 2018.
\newblock The {N}arrative{QA} reading comprehension challenge.
\newblock \emph{TACL}, 6:317--328.

\bibitem[{Lai et~al.(2017)Lai, Xie, Liu, Yang, and Hovy}]{lai2017race}
Guokun Lai, Qizhe Xie, Hanxiao Liu, Yiming Yang, and Eduard Hovy. 2017.
\newblock \href {https://doi.org/10.18653/v1/D17-1082} {{RACE}: Large-scale
  {R}e{A}ding comprehension dataset from examinations}.
\newblock In \emph{Proceedings of the 2017 Conference on Empirical Methods in
  Natural Language Processing}, pages 785--794, Copenhagen, Denmark.
  Association for Computational Linguistics.

\bibitem[{Lan et~al.(2020)Lan, Chen, Goodman, Gimpel, Sharma, and
  Soricut}]{lan2019albert}
Zhenzhong Lan, Mingda Chen, Sebastian Goodman, Kevin Gimpel, Piyush Sharma, and
  Radu Soricut. 2020.
\newblock \href {https://openreview.net/forum?id=H1eA7AEtvS} {{ALBERT:} {A}
  lite {BERT} for self-supervised learning of language representations}.
\newblock In \emph{8th International Conference on Learning Representations,
  {ICLR} 2020, Addis Ababa, Ethiopia, April 26-30, 2020}. OpenReview.net.

\bibitem[{Lewis et~al.(2020)Lewis, Liu, Goyal, Ghazvininejad, Mohamed, Levy,
  Stoyanov, and Zettlemoyer}]{lewis2020bart}
Mike Lewis, Yinhan Liu, Naman Goyal, Marjan Ghazvininejad, Abdelrahman Mohamed,
  Omer Levy, Veselin Stoyanov, and Luke Zettlemoyer. 2020.
\newblock \href {https://doi.org/10.18653/v1/2020.acl-main.703} {{BART}:
  Denoising sequence-to-sequence pre-training for natural language generation,
  translation, and comprehension}.
\newblock In \emph{Proceedings of the 58th Annual Meeting of the Association
  for Computational Linguistics}, pages 7871--7880, Online. Association for
  Computational Linguistics.

\bibitem[{Li et~al.(2018)Li, He, Cai, Zhang, Zhao, Liu, Li, and Si}]{li2018srl}
Zuchao Li, Shexia He, Jiaxun Cai, Zhuosheng Zhang, Hai Zhao, Gongshen Liu,
  Linlin Li, and Luo Si. 2018.
\newblock \href {https://doi.org/10.18653/v1/D18-1262} {A unified syntax-aware
  framework for semantic role labeling}.
\newblock In \emph{Proceedings of the 2018 Conference on Empirical Methods in
  Natural Language Processing}, pages 2401--2411, Brussels, Belgium.
  Association for Computational Linguistics.

\bibitem[{Liu et~al.(2017)Liu, Hu, Wei, Yang, and Nyberg}]{liu2017structural}
Rui Liu, Junjie Hu, Wei Wei, Zi~Yang, and Eric Nyberg. 2017.
\newblock \href {https://doi.org/10.18653/v1/D17-1085} {Structural embedding of
  syntactic trees for machine comprehension}.
\newblock In \emph{Proceedings of the 2017 Conference on Empirical Methods in
  Natural Language Processing}, pages 815--824, Copenhagen, Denmark.
  Association for Computational Linguistics.

\bibitem[{Liu et~al.(2019)Liu, Ott, Goyal, Du, Joshi, Chen, Levy, Lewis,
  Zettlemoyer, and Stoyanov}]{liu2019roberta}
Yinhan Liu, Myle Ott, Naman Goyal, Jingfei Du, Mandar Joshi, Danqi Chen, Omer
  Levy, Mike Lewis, Luke Zettlemoyer, and Veselin Stoyanov. 2019.
\newblock Roberta: A robustly optimized bert pretraining approach.
\newblock \emph{arXiv preprint arXiv:1907.11692}.

\bibitem[{Mihaylov and Frank(2019)}]{mihaylov2019discourse}
Todor Mihaylov and Anette Frank. 2019.
\newblock \href {https://doi.org/10.18653/v1/D19-1257} {Discourse-aware
  semantic self-attention for narrative reading comprehension}.
\newblock In \emph{Proceedings of the 2019 Conference on Empirical Methods in
  Natural Language Processing and the 9th International Joint Conference on
  Natural Language Processing (EMNLP-IJCNLP)}, pages 2541--2552, Hong Kong,
  China. Association for Computational Linguistics.

\bibitem[{Paszke et~al.(2019)Paszke, Gross, Massa, Lerer, Bradbury, Chanan,
  Killeen, Lin, Gimelshein, Antiga, Desmaison, Kopf, Yang, DeVito, Raison,
  Tejani, Chilamkurthy, Steiner, Fang, Bai, and Chintala}]{paszke2019pytorch}
Adam Paszke, Sam Gross, Francisco Massa, Adam Lerer, James Bradbury, Gregory
  Chanan, Trevor Killeen, Zeming Lin, Natalia Gimelshein, Luca Antiga, Alban
  Desmaison, Andreas Kopf, Edward Yang, Zachary DeVito, Martin Raison, Alykhan
  Tejani, Sasank Chilamkurthy, Benoit Steiner, Lu~Fang, Junjie Bai, and Soumith
  Chintala. 2019.
\newblock \href
  {http://papers.neurips.cc/paper/9015-pytorch-an-imperative-style-high-performance-deep-learning-library.pdf}
  {Pytorch: An imperative style, high-performance deep learning library}.
\newblock In \emph{Advances in Neural Information Processing Systems 32}, pages
  8024--8035. Curran Associates, Inc.

\bibitem[{Raffel and Ellis(2015)}]{Raffel2015self}
Colin Raffel and Daniel P.~W. Ellis. 2015.
\newblock \href {http://arxiv.org/abs/1512.08756} {Feed-forward networks with
  attention can solve some long-term memory problems}.

\bibitem[{Raffel et~al.(2020)Raffel, Shazeer, Roberts, Lee, Narang, Matena,
  Zhou, Li, and Liu}]{raffel2020t5}
Colin Raffel, Noam Shazeer, Adam Roberts, Katherine Lee, Sharan Narang, Michael
  Matena, Yanqi Zhou, Wei Li, and Peter~J. Liu. 2020.
\newblock Exploring the limits of transfer learning with a unified text-to-text
  transformer.
\newblock \emph{JMLR}.

\bibitem[{Rajpurkar et~al.(2018)Rajpurkar, Jia, and
  Liang}]{rajpurkar2018squad2}
Pranav Rajpurkar, Robin Jia, and Percy Liang. 2018.
\newblock \href {https://doi.org/10.18653/v1/P18-2124} {Know what you don{'}t
  know: Unanswerable questions for {SQ}u{AD}}.
\newblock In \emph{Proceedings of the 56th Annual Meeting of the Association
  for Computational Linguistics (Volume 2: Short Papers)}, pages 784--789,
  Melbourne, Australia. Association for Computational Linguistics.

\bibitem[{Rajpurkar et~al.(2016)Rajpurkar, Zhang, Lopyrev, and
  Liang}]{rajpurkar2016squad}
Pranav Rajpurkar, Jian Zhang, Konstantin Lopyrev, and Percy Liang. 2016.
\newblock \href {https://doi.org/10.18653/v1/D16-1264} {{SQ}u{AD}: 100,000+
  questions for machine comprehension of text}.
\newblock In \emph{Proceedings of the 2016 Conference on Empirical Methods in
  Natural Language Processing}, pages 2383--2392, Austin, Texas. Association
  for Computational Linguistics.

\bibitem[{Seo et~al.(2017)Seo, Kembhavi, Farhadi, and
  Hajishirzi}]{seo2017bidaf}
Minjoon Seo, Aniruddha Kembhavi, Ali Farhadi, and Hannaneh Hajishirzi. 2017.
\newblock \href {https://openreview.net/forum?id=HJ0UKP9ge} {Bidirectional
  attention flow for machine comprehension}.
\newblock In \emph{International Conference on Learning Representations}.

\bibitem[{Shoeybi et~al.(2019)Shoeybi, Patwary, Puri, LeGresley, Casper, and
  Catanzaro}]{shoeybi2019megatron}
Mohammad Shoeybi, Mostofa Patwary, Raul Puri, Patrick LeGresley, Jared Casper,
  and Bryan Catanzaro. 2019.
\newblock \href {http://arxiv.org/abs/1909.08053} {Megatron-lm: Training
  multi-billion parameter language models using model parallelism}.

\bibitem[{Strubell et~al.(2018)Strubell, Verga, Andor, Weiss, and
  McCallum}]{Strubell2018linguistically}
Emma Strubell, Patrick Verga, Daniel Andor, David Weiss, and Andrew McCallum.
  2018.
\newblock \href {https://doi.org/10.18653/v1/D18-1548} {Linguistically-informed
  self-attention for semantic role labeling}.
\newblock In \emph{Proceedings of the 2018 Conference on Empirical Methods in
  Natural Language Processing}, pages 5027--5038, Brussels, Belgium.
  Association for Computational Linguistics.

\bibitem[{Sun et~al.(2019{\natexlab{a}})Sun, Yu, Chen, Yu, Choi, and
  Cardie}]{sun2019dream}
Kai Sun, Dian Yu, Jianshu Chen, Dong Yu, Yejin Choi, and Claire Cardie.
  2019{\natexlab{a}}.
\newblock \href {https://doi.org/10.1162/tacl_a_00264} {{DREAM}: A challenge
  data set and models for dialogue-based reading comprehension}.
\newblock \emph{Transactions of the Association for Computational Linguistics},
  7:217--231.

\bibitem[{Sun et~al.(2019{\natexlab{b}})Sun, Yu, Yu, and
  Cardie}]{sun2019strategy}
Kai Sun, Dian Yu, Dong Yu, and Claire Cardie. 2019{\natexlab{b}}.
\newblock \href {https://doi.org/10.18653/v1/N19-1270} {Improving machine
  reading comprehension with general reading strategies}.
\newblock In \emph{Proceedings of the 2019 Conference of the North {A}merican
  Chapter of the Association for Computational Linguistics: Human Language
  Technologies, Volume 1 (Long and Short Papers)}, pages 2633--2643,
  Minneapolis, Minnesota. Association for Computational Linguistics.

\bibitem[{Trischler et~al.(2017)Trischler, Wang, Yuan, Harris, Sordoni,
  Bachman, and Suleman}]{trischler2017newsqa}
Adam Trischler, Tong Wang, Xingdi Yuan, Justin Harris, Alessandro Sordoni,
  Philip Bachman, and Kaheer Suleman. 2017.
\newblock \href {https://doi.org/10.18653/v1/W17-2623} {{N}ews{QA}: A machine
  comprehension dataset}.
\newblock In \emph{Proceedings of the 2nd Workshop on Representation Learning
  for {NLP}}, pages 191--200, Vancouver, Canada. Association for Computational
  Linguistics.

\bibitem[{Vaswani et~al.(2017)Vaswani, Shazeer, Parmar, Uszkoreit, Jones,
  Gomez, Kaiser, and Polosukhin}]{vaswani2017attention}
Ashish Vaswani, Noam Shazeer, Niki Parmar, Jakob Uszkoreit, Llion Jones,
  Aidan~N Gomez, \L~ukasz Kaiser, and Illia Polosukhin. 2017.
\newblock Attention is all you need.
\newblock In \emph{NIPS}, volume~30.

\bibitem[{Wang et~al.(2017)Wang, Yang, Wei, Chang, and Zhou}]{wang2017gated}
Wenhui Wang, Nan Yang, Furu Wei, Baobao Chang, and Ming Zhou. 2017.
\newblock \href {https://doi.org/10.18653/v1/P17-1018} {Gated self-matching
  networks for reading comprehension and question answering}.
\newblock In \emph{Proceedings of the 55th Annual Meeting of the Association
  for Computational Linguistics (Volume 1: Long Papers)}, pages 189--198,
  Vancouver, Canada. Association for Computational Linguistics.

\bibitem[{Xiong et~al.(2017)Xiong, Zhong, and Socher}]{xiong2017coattention}
Caiming Xiong, Victor Zhong, and Richard Socher. 2017.
\newblock Dynamic coattention networks for question answering.
\newblock In \emph{ICLR}.

\bibitem[{Yamada et~al.(2020)Yamada, Asai, Shindo, Takeda, and
  Matsumoto}]{yamada2020luke}
Ikuya Yamada, Akari Asai, Hiroyuki Shindo, Hideaki Takeda, and Yuji Matsumoto.
  2020.
\newblock \href {https://doi.org/10.18653/v1/2020.emnlp-main.523} {{LUKE}: Deep
  contextualized entity representations with entity-aware self-attention}.
\newblock In \emph{Proceedings of the 2020 Conference on Empirical Methods in
  Natural Language Processing (EMNLP)}, pages 6442--6454, Online. Association
  for Computational Linguistics.

\bibitem[{Yang et~al.(2019)Yang, Dai, Yang, Carbonell, Salakhutdinov, and
  Le}]{yang2019xlnet}
Zhilin Yang, Zihang Dai, Yiming Yang, Jaime~G. Carbonell, Ruslan Salakhutdinov,
  and Quoc~V. Le. 2019.
\newblock \href
  {https://proceedings.neurips.cc/paper/2019/hash/dc6a7e655d7e5840e66733e9ee67cc69-Abstract.html}
  {Xlnet: Generalized autoregressive pretraining for language understanding}.
\newblock In \emph{Advances in Neural Information Processing Systems 32: Annual
  Conference on Neural Information Processing Systems 2019, NeurIPS 2019,
  December 8-14, 2019, Vancouver, BC, Canada}, pages 5754--5764.

\bibitem[{Yang et~al.(2018)Yang, Qi, Zhang, Bengio, Cohen, Salakhutdinov, and
  Manning}]{yang2018hotpotqa}
Zhilin Yang, Peng Qi, Saizheng Zhang, Yoshua Bengio, William Cohen, Ruslan
  Salakhutdinov, and Christopher~D. Manning. 2018.
\newblock \href {https://doi.org/10.18653/v1/D18-1259} {{H}otpot{QA}: A dataset
  for diverse, explainable multi-hop question answering}.
\newblock In \emph{Proceedings of the 2018 Conference on Empirical Methods in
  Natural Language Processing}, pages 2369--2380, Brussels, Belgium.
  Association for Computational Linguistics.

\bibitem[{Yu et~al.(2018)Yu, Dohan, Le, Luong, Zhao, and Chen}]{wei2018qanet}
Adams~Wei Yu, David Dohan, Quoc Le, Thang Luong, Rui Zhao, and Kai Chen. 2018.
\newblock \href {https://openreview.net/forum?id=B14TlG-RW} {Qanet: Combining
  local convolution with global self-attention for reading comprehension}.
\newblock In \emph{International Conference on Learning Representations}.

\bibitem[{Zhang et~al.(2020{\natexlab{a}})Zhang, Zhao, Wu, Zhang, Zhou, and
  Zhou}]{zhang2020dcmn}
Shuailiang Zhang, Hai Zhao, Yuwei Wu, Zhuosheng Zhang, Xi~Zhou, and Xiang Zhou.
  2020{\natexlab{a}}.
\newblock {DCMN}+: Dual co-matching network for multi-choice reading
  comprehension.
\newblock In \emph{AAAI}.

\bibitem[{Zhang et~al.(2020{\natexlab{b}})Zhang, Wu, Zhao, Li, Zhang, Zhou, and
  Zhou}]{zhang2020semBERT}
Zhuosheng Zhang, Yuwei Wu, Hai Zhao, Zuchao Li, Shuailiang Zhang, Xi~Zhou, and
  Xiang Zhou. 2020{\natexlab{b}}.
\newblock \href {https://aaai.org/ojs/index.php/AAAI/article/view/6510}
  {Semantics-aware {BERT} for language understanding}.
\newblock In \emph{The Thirty-Fourth {AAAI} Conference on Artificial
  Intelligence, {AAAI} 2020, The Thirty-Second Innovative Applications of
  Artificial Intelligence Conference, {IAAI} 2020, The Tenth {AAAI} Symposium
  on Educational Advances in Artificial Intelligence, {EAAI} 2020, New York,
  NY, USA, February 7-12, 2020}, pages 9628--9635. {AAAI} Press.

\bibitem[{Zhang et~al.(2020{\natexlab{c}})Zhang, Wu, Zhou, Duan, Zhao, and
  Wang}]{zhang2020sgnet}
Zhuosheng Zhang, Yuwei Wu, Junru Zhou, Sufeng Duan, Hai Zhao, and Rui Wang.
  2020{\natexlab{c}}.
\newblock Sg-net: Syntax-guided machine reading comprehension.
\newblock \emph{AAAI}.

\bibitem[{Zhang et~al.(2021)Zhang, Yang, and Zhao}]{zhang2020retro}
Zhuosheng Zhang, Junjie Yang, and Hai Zhao. 2021.
\newblock Retrospective reader for machine reading comprehension.
\newblock In \emph{AAAI}.

\bibitem[{Zheng et~al.(2020)Zheng, Wen, Liang, Duan, Che, Jiang, Zhou, and
  Liu}]{zheng2020document}
Bo~Zheng, Haoyang Wen, Yaobo Liang, Nan Duan, Wanxiang Che, Daxin Jiang, Ming
  Zhou, and Ting Liu. 2020.
\newblock \href {https://doi.org/10.18653/v1/2020.acl-main.599} {Document
  modeling with graph attention networks for multi-grained machine reading
  comprehension}.
\newblock In \emph{Proceedings of the 58th Annual Meeting of the Association
  for Computational Linguistics}, pages 6708--6718, Online. Association for
  Computational Linguistics.

\bibitem[{Zhou et~al.(2020)Zhou, Zhang, Zhao, and Zhang}]{zhou2020limit}
Junru Zhou, Zhuosheng Zhang, Hai Zhao, and Shuailiang Zhang. 2020.
\newblock \href {https://doi.org/10.18653/v1/2020.findings-emnlp.399}
  {{LIMIT}-{BERT} : Linguistics informed multi-task {BERT}}.
\newblock In \emph{Findings of the Association for Computational Linguistics:
  EMNLP 2020}, pages 4450--4461, Online. Association for Computational
  Linguistics.

\bibitem[{Zhou and Zhao(2019)}]{zhou2019head}
Junru Zhou and Hai Zhao. 2019.
\newblock \href {https://doi.org/10.18653/v1/P19-1230} {{H}ead-{D}riven
  {P}hrase {S}tructure {G}rammar parsing on {P}enn {T}reebank}.
\newblock In \emph{Proceedings of the 57th Annual Meeting of the Association
  for Computational Linguistics}, pages 2396--2408, Florence, Italy.
  Association for Computational Linguistics.

\bibitem[{Zhu et~al.(2020)Zhu, Zhao, and Li}]{zhu2020dual}
Pengfei Zhu, Hai Zhao, and Xiaoguang Li. 2020.
\newblock Dual multi-head co-attention for multi-choice reading comprehension.
\newblock \emph{arXiv preprint arXiv:2001.09415}.

\end{thebibliography}
\bibliographystyle{acl_natbib}

\newpage
\appendix
\section{Part-Of-Speech Tags List}
\label{Part-Of-Speech Tags List}
In this appendix, we list all 39 POS tags (including POS tags from \emph{nltk} POS tagger and defined by us) in Table \ref{tags}.

\section{Complete Comparison Results on Benchmarks}
\label{Complete Comparison Results on Benchmarks}
We show complete public works on DREAM, RACE, SQuAD 1.1 and SQuAD in this appendix, as Tables \ref{dream_result} \ref{race_result}, \ref{squad1.1_result} and \ref{squad2.0_result} show.

The results show that, our \emph{POI-Net} outperforms most of comparison models and baselines, expect models:
1) with massive and incomparable parameters like T5 \cite{raffel2020t5} and Megatron-BERT \cite{shoeybi2019megatron};
2) in more advanced baseline architecture like XLNet \cite{yang2019xlnet}, ELECTRA \cite{clark2019electra};
3) in special model design for one single subcategory of discriminative MRC task \cite{zhang2020retro}.

\begin{table}[ht]
\centering{
	\begin{tabular}{p{5.3cm}|c|c}
		\hline\bf Model & \bf Dev & \bf Test \\\hline\hline
		FTLM++ \cite{sun2019dream} & 58.1 & 58.2    \\
		BERT$_{base}$ \cite{devlin2019bert} & 63.4 & 63.2  \\
		BERT$_{large}$ \cite{devlin2019bert} & 66.0 & 66.8   \\
		XLNet$_{large}$ \cite{yang2019xlnet} & --  & 72.0 \\
		RoBERTa$_{large}$ \cite{liu2019roberta} & 85.4 & 85.0  \\
		RoBERTa$_{large}$ + MMM \cite{jin2019mmm} & 88.0 & 88.9 \\
		ALBERT$_{xxlarge}$ + DUMA \cite{zhu2020dual} & 89.9 & 90.4 \\
		ALBERT$_{xxlarge}$ + DUMA + MTL & -- & 91.8 \\\hline
		ALBERT$_{base}$ (rerun) & 65.7 & 65.6 \\
		POI-Net on ALBERT$_{base}$ & 68.6 & 68.5 \\\hline
		ALBERT$_{xxlarge}$ (rerun) & 89.2 & 88.5 \\
		POI-Net on ALBERT$_{xxlarge}$ & 90.0 & 90.3 \\\hline\hline
	\end{tabular}
	\caption{\label{dream_result} Public submissions on DREAM. The results in the first domain are from the leaderboard. MTL denotes multi-task learning.}
	}
\end{table}

\begin{table}[ht]
	\centering{
		\begin{tabular}{l|p{5.5cm}}
		\hline\bf POS Tag & \bf Meaning \\\hline\hline
        CC & Coordinating conjunction\\
        CD & Cardinal number\\
        DT & Determiner\\
        EX & Existential there\\
        FW & Foreign word\\
        IN & Preposition or subordinating conjunction\\
        JJ & Adjective\\
        JJR & Adjective, comparative\\
        JJS & Adjective, superlative\\
        LS & List item marker\\
        MD & Modal\\
        NN & Noun, singular or mass\\
        NNS & Noun, plural\\
        NNP & Proper noun, singular\\
        NNPS & Proper noun, plural\\
        PDT & Predeterminer\\
        POS & Possessive ending\\
        PRP & Personal pronoun\\
        PRP\$ & Possessive pronoun\\
        RB & Adverb\\
        RBR & Adverb, comparative\\
        RBS & Adverb, superlative\\
        RP & Particle\\
        SYM & Symbol\\
        TO & To\\
        UH & Interjection\\
        VB & Verb, base form\\
        VBD & Verb, past tense\\
        VBG & Verb, gerund or present participle\\
        VBN & Verb, past participle\\
        VBP & Verb, non-3rd person singular present\\
        VBZ & Verb, 3rd person singular present\\
        WDT & Wh-determiner\\
        WP &  Wh-pronoun\\
        WP\$ & Possessive wh-pronoun\\
        WRB & Wh-adverb\\
        SPE & Special tokens: [CLS], [SEP]\\
        PAD & Padding tokens\\
        ERR & Unrecognized tokens\\
        \hline
		\end{tabular}
	}
	\caption{\label{tags} The complete list for all POS tags in \emph{POI-Net}.}
\end{table}

\begin{table*}[ht]
\centering
{
	\begin{tabular}{l|c|c}
		\hline\bf Model & \bf Dev (M / H) & \bf Test (M / H)\\\hline\hline
		BERT$_{base}$ \cite{devlin2019bert} & 64.6 (-- / --) & 65.0 (71.1 / 62.3)  \\
		BERT$_{large}$ \cite{devlin2019bert} & 72.7 (76.7 / 71.0) & 72.0 (76.6 / 70.1)  \\
		XLNet$_{large}$ \cite{yang2019xlnet} & 80.1 (-- / --) & 81.8 (85.5 / 80.2)\\
		XLNet$_{large}$ + DCMN+ \cite{zhang2020dcmn} & -- (-- / --) & 82.8 (86.5 / 81.3)\\
		RoBERTa$_{large}$ \cite{liu2019roberta} & -- (-- / --) & 83.2 (86.5 / 81.8)\\
		RoBERTa$_{large}$ + MMM \cite{jin2019mmm} & -- (-- / --) & 85.0 (89.1 / 83.3)\\
		T5-11B \cite{raffel2020t5} & -- (-- / --) & 87.1 (-- / --)\\
		ALBERT$_{xxlarge}$ + DUMA \cite{zhu2020dual} & 88.1 (-- / --) & 88.0 (90.9 / 86.7)\\
		T5-11B + UnifiedQA \cite{khashabi2020unifiedqa} & -- (-- / --) & 89.4 (-- / --)\\
		Megatron-BERT-3.9B \cite{shoeybi2019megatron} & -- (-- / --) & 89.5 (91.8 / 88.6)\\
		ALBERT$_{xxlarge}$ + SC + TL \cite{jiang2020single} & -- (-- / --) & 90.7 (92.8 / 89.8) \\\hline
        ALBERT$_{base}$ (rerun) & 67.9 (72.3 / 65.7) & 67.2 (72.1 / 65.2) \\
		POI-Net on ALBERT$_{base}$  & 72.4 (76.3 / 70.0) & 71.0 (75.7 / 69.0) \\\hline
        ALBERT$_{xxlarge}$ (rerun) & 86.6 (89.4 / 85.2) & 86.5 (89.2 / 85.4) \\
		POI-Net on ALBERT$_{xxlarge}$ & 88.1 (91.3 / 86.3) & 88.3 (91.5 / 86.8)\\\hline\hline
	\end{tabular}
	\caption{\label{race_result} Public submissions on RACE. The results in the first domain are from the leaderboard. SC denotes single choice and TL denotes transfer learning.}
	}
\end{table*}

\begin{table}[ht]
\centering{
	\begin{tabular}{p{5.3cm}|c|c}
		\hline\bf Model & \bf EM & \bf F1 \\\hline\hline
		SAN \cite{liu2017structural} & 76.2 & 84.1  \\
		R.M-Reader \cite{hu2018rm} & 81.2 & 87.9  \\
		ALBERT$_{base}$ \cite{lan2019albert} & 82.9 & 89.3  \\
		BERT$_{base}$ \cite{devlin2019bert} & 80.8 & 88.5  \\
		BERT$_{large}$ \cite{devlin2019bert} & 85.5 & 92.2   \\
		ALBERT$_{xxlarge}$ \cite{lan2019albert} & 88.3 & 94.1 \\
		SpanBERT$^*$ \cite{joshi2020spanbert} & 88.8 & 94.6 \\
		XLNet$_{large}$ \cite{yang2019xlnet} & 89.7  & 95.1 \\
		RoBERTa$_{large}$ + LUKE \cite{yamada2020luke} & 89.8 & 95.0  \\
		\hline
		ALBERT$_{base}$ (rerun) & 82.7 & 89.9 \\
		POI-Net on ALBERT$_{base}$ & 84.5 & 91.3 \\\hline
		ALBERT$_{xxlarge}$ (rerun) & 88.2 & 94.1 \\
		POI-Net on ALBERT$_{xxlarge}$ & 89.5 & 95.0 \\\hline\hline
	\end{tabular}
	\caption{\label{squad1.1_result} Comparison works on SQuAD 1.1 development set. Results with $^*$ are from \cite{clark2019electra}.}
	}
\end{table}

\begin{table}[ht]
\centering{
	\begin{tabular}{p{5.3cm}|c|c}
		\hline\bf Model & \bf EM & \bf F1 \\\hline\hline
		ALBERT$_{base}$ \cite{lan2019albert} & 77.1 & 80.0  \\
		BERT$_{base}$ \cite{devlin2019bert} & 77.6 & 80.4  \\
		NeurQuRI \cite{back2020NeurQuRI} & 80.0 & 83.1 \\
		BERT$_{large}$ \cite{devlin2019bert} & 82.2 & 85.0   \\
		SemBERT \cite{zhang2020semBERT} & 84.2 & 87.9 \\
		ALBERT$_{xxlarge}$ \cite{lan2019albert} & 85.1 & 88.1 \\
		SpanBERT$^*$ \cite{joshi2020spanbert} & 85.7 & 88.7 \\
		XLNet$_{large}$ \cite{yang2019xlnet} & 87.9 & 90.6 \\
		ELECTRA \cite{clark2019electra} & 88.0 & 90.6 \\
		ALBERT$_{xxlarge}$ + Retro-Reader \cite{zhang2020retro} & 87.8 & 90.9 \\
		\hline
		ALBERT$_{base}$ (rerun) & 77.3 & 80.4 \\
		POI-Net on ALBERT$_{base}$ & 79.8 & 82.9 \\\hline
		ALBERT$_{xxlarge}$ (rerun) & 85.4 & 88.5 \\
		POI-Net on ALBERT$_{xxlarge}$ & 87.7 & 90.6 \\\hline\hline
	\end{tabular}
	\caption{\label{squad2.0_result} Comparison works on SQuAD 2.0 development set. Results with $^*$ are from \cite{clark2019electra}.}
	}
\end{table}

\end{document}